\documentclass[11pt]{article}

\usepackage[preprint]{acl}

\usepackage{times}
\usepackage{latexsym}

\usepackage[T1]{fontenc}

\usepackage[utf8]{inputenc}

\usepackage{microtype}

\usepackage{inconsolata}

\usepackage{graphicx}
\usepackage{amsmath}
\usepackage{amssymb}
\usepackage{subcaption}
\usepackage[table]{xcolor}
\usepackage{booktabs}   
\usepackage{xcolor}     
\usepackage{arydshln}   
\usepackage{caption}   
\usepackage{adjustbox} 

\title{
    \begin{minipage}{0.12\textwidth} 
        \raggedleft
        \includegraphics[height=1.5cm]{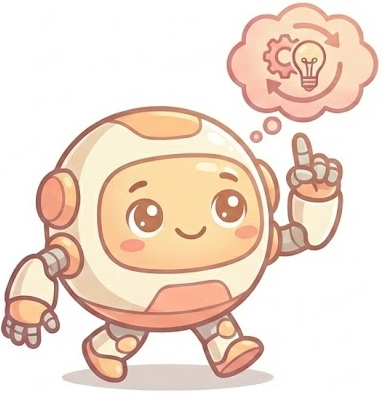} 
    \end{minipage}%
    \hspace{0.005cm} 
    \begin{minipage}{0.8\textwidth} 
        \centering
        \textsc{iReasoner}: Trajectory-Aware Intrinsic Reasoning Supervision for Self-Evolving Large Multimodal Models
    \end{minipage}
}

\author{
  \textbf{Meghana Sunil\textsuperscript{1,2}\thanks{Correspondence: \href{mailto:meghanaa.sunil@gmail.com}{meghanaa.sunil@gmail.com}, \href{mailto:kavitha@nagasaki-u.ac.jp}{kavitha@nagasaki-u.ac.jp}}},
  \textbf{Manikandarajan Venmathimaran\textsuperscript{3}},
  \textbf{Muthu Subash Kavitha\textsuperscript{2}}
  \\
  \\
  \textsuperscript{1}Vellore Institute of Technology, Chennai, India \\
  \textsuperscript{2}School of Information and Data Sciences, Nagasaki University, Nagasaki, Japan \\
  \textsuperscript{3}Loughborough University, United Kingdom
}

\begin{document}
\maketitle
\begin{abstract}
Recent work shows that large multimodal models (LMMs) can self-improve from unlabeled data via self-play and intrinsic feedback. Yet existing self-evolving frameworks mainly reward final outcomes, leaving intermediate reasoning weakly constrained despite its importance for visually grounded decision making. We propose \textsc{iReasoner}, a self-evolving framework that improves an LMM's \emph{implicit} reasoning by explicitly eliciting chain-of-thought (CoT) and rewarding its internal agreement. In a \emph{Proposer}--\emph{Solver} loop over unlabeled images, \textsc{iReasoner} augments outcome-level intrinsic rewards with a trajectory-aware signal defined over intermediate reasoning steps, providing learning signals that distinguish reasoning paths leading to the same answer without ground-truth labels or external judges. Starting from Qwen2.5-VL-7B, \textsc{iReasoner} yields up to $+2.1$ points across diverse multimodal reasoning benchmarks under fully unsupervised post-training. We hope this work serves as a starting point for reasoning-aware self-improvement in LMMs in purely unsupervised settings. Our code is available \textcolor{blue}{\href{https://meghanaasunil.github.io/iReasoner/}{here}}.

\end{abstract}

\section{Introduction}

Self-improvement has become a practical way to push foundation models beyond supervised instruction tuning by generating training experiences and learning from internal feedback~\citep{SelfImprovementSurvey,intro1,intro2,intro3,intro4}. In large multimodal models (LMMs), this idea has recently enabled self-evolving pipelines that train directly on streams of unlabeled images: a model proposes visually grounded questions, samples multiple solution attempts, and updates itself with intrinsic rewards computed from its own outputs~\cite{thawakar2025evolmmselfevolvinglargemultimodal,he2025visplayselfevolvingvisionlanguagemodels}. 
These approaches point to an appealing scaling path, since raw visual data is abundant while high-quality multimodal annotations and external reward systems remain costly to obtain~\citep{intro5,intro6}.

\begin{figure}[!t]
  \centering
  \includegraphics[width=\columnwidth]{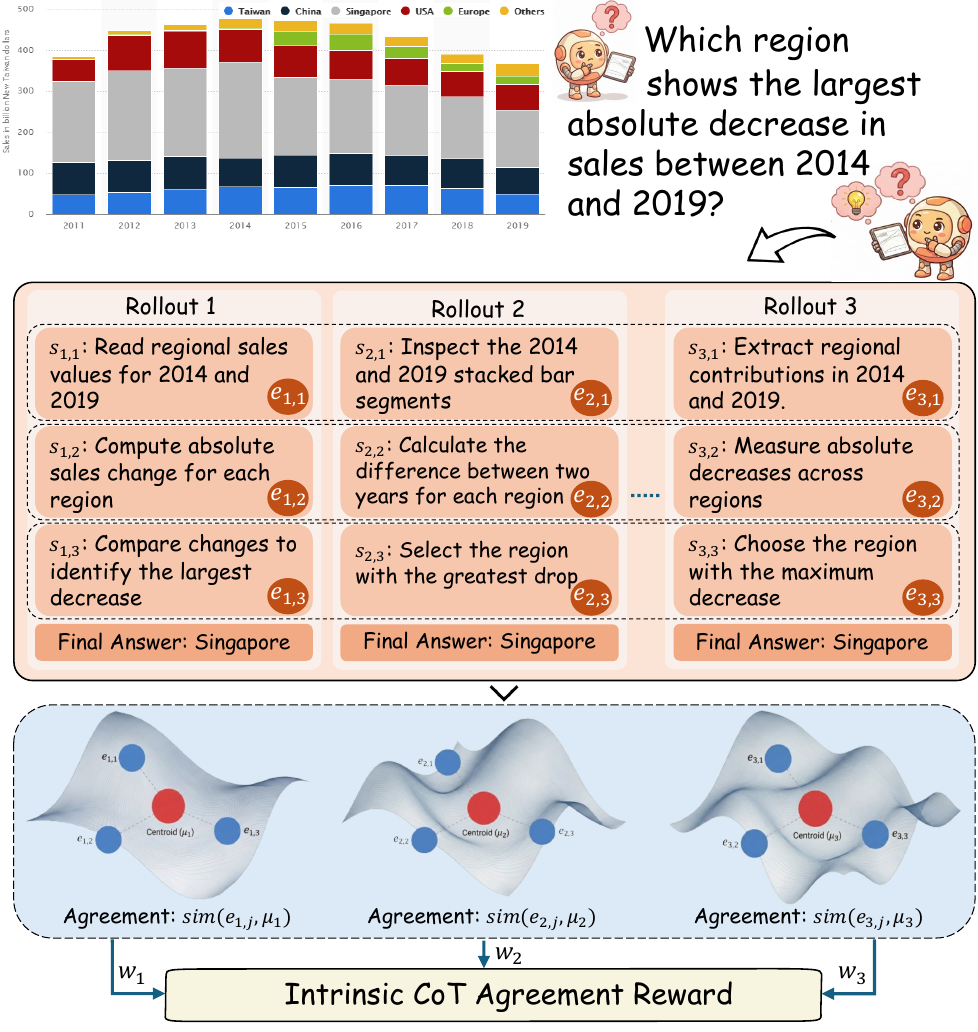}
  \caption{\textbf{\textsc{iReasoner}'s intrinsic step-level CoT agreement.}
  Given an unlabeled image, the \emph{Proposer} generates a visually grounded question, and the \emph{Solver} samples $N$ reasoning rollouts, each producing a CoT with multiple intermediate steps and a final answer (3 rollouts and 3 steps are shown here).
  Among rollouts in the dominant (majority-answer) group, we embed each step text $s_{i,j}$ into $e_{i,j}$ and form a per-step prototype $\mu_j$.
  Step agreement is computed via similarity $\mathrm{sim}(e_{i,j},\mu_j)$ and aggregated with higher weight on earlier, grounding-heavy steps ($w_1>w_2>w_3$) to produce a scalar Intrinsic CoT Agreement Reward.}
  \label{fig:intrinsic-cot-agreement-qual}
\end{figure}

A central limitation is that most self-improving LMM frameworks still define verification over final outcomes---such as a final answer, a whole response, a preference score, or a reconstruction score---while leaving intermediate reasoning weakly constrained~\cite{l1p9,l1p10,SelfImprovementSurvey}. 
This matters even when the goal is not to build an explicit ``reasoning model'': instruction-tuned LMMs often rely on implicit reasoning, and we can elicit step-by-step chains of thought (CoT) during training to shape that latent computation~\citep{intro7,intro8,intro9,intro10}.
Under outcome-only rewards, two responses that reach the same answer receive nearly the same learning signal, even if one is grounded in the image and the other relies on shortcuts or hallucinated intermediate claims that happen to cancel out.
Outcome-only optimization therefore provides little direct pressure to stabilize the sequence of intermediate claims that drives reliable vision--language reasoning.

Prior work has made steady progress on self-improvement, but step-aware signals are still difficult to obtain in a purely unsupervised setting.
Early approaches were developed mainly for large language models (LLMs)~\cite{llm1,llm2,llm3} and later extended to LMMs using labeled supervision, external judges, or auxiliary reward systems~\cite{l1p5,l1p6,l1p7}. 
Other approaches sidestep step supervision by using coarse consistency or reconstruction objectives to curate synthetic training data~\cite{l1p9,l1p10}. 
More recently, EvoLMM~\cite{thawakar2025evolmmselfevolvinglargemultimodal} and VisPlay~\cite{he2025visplayselfevolvingvisionlanguagemodels} showed that purely unlabeled self-evolution from raw images is feasible; however, their intrinsic rewards still operate primarily at the answer/response level, leaving open the problem of how to evaluate and optimize CoT steps in multimodal self-evolution without labeled data or external supervision.

We address this gap by introducing \textsc{iReasoner}, a self-evolving framework that improves an LMM's \emph{implicit} reasoning by eliciting explicit step-by-step CoT and rewarding its internal agreement.
We follow the \emph{Proposer}--\emph{Solver} self-evolution backbone on unlabeled images~\cite{thawakar2025evolmmselfevolvinglargemultimodal,he2025visplayselfevolvingvisionlanguagemodels}, but differ in how the \emph{Solver} is supervised: unlike prior methods whose intrinsic rewards are defined at the answer/response level, \textsc{iReasoner} introduces a trajectory-aware reward defined over intermediate CoT steps.
As illustrated in Fig.~\ref{fig:intrinsic-cot-agreement-qual}, we group \emph{Solver} rollouts by the dominant (majority) answer, align their CoT into step indices, and compute per-step prototypes from the dominant group.
Each rollout is then rewarded by its step-wise similarity to these prototypes, with higher weight on early, grounding-heavy steps, yielding a single scalar Intrinsic CoT Agreement Reward.
This reward is fully intrinsic and can drive policy-gradient updates without ground-truth labels, external judges, or verifiers, while distinguishing between reasoning trajectories that arrive at the same final answer but differ in intermediate claims.

\begin{figure*}[t]
  \centering
  \includegraphics[width=\textwidth]{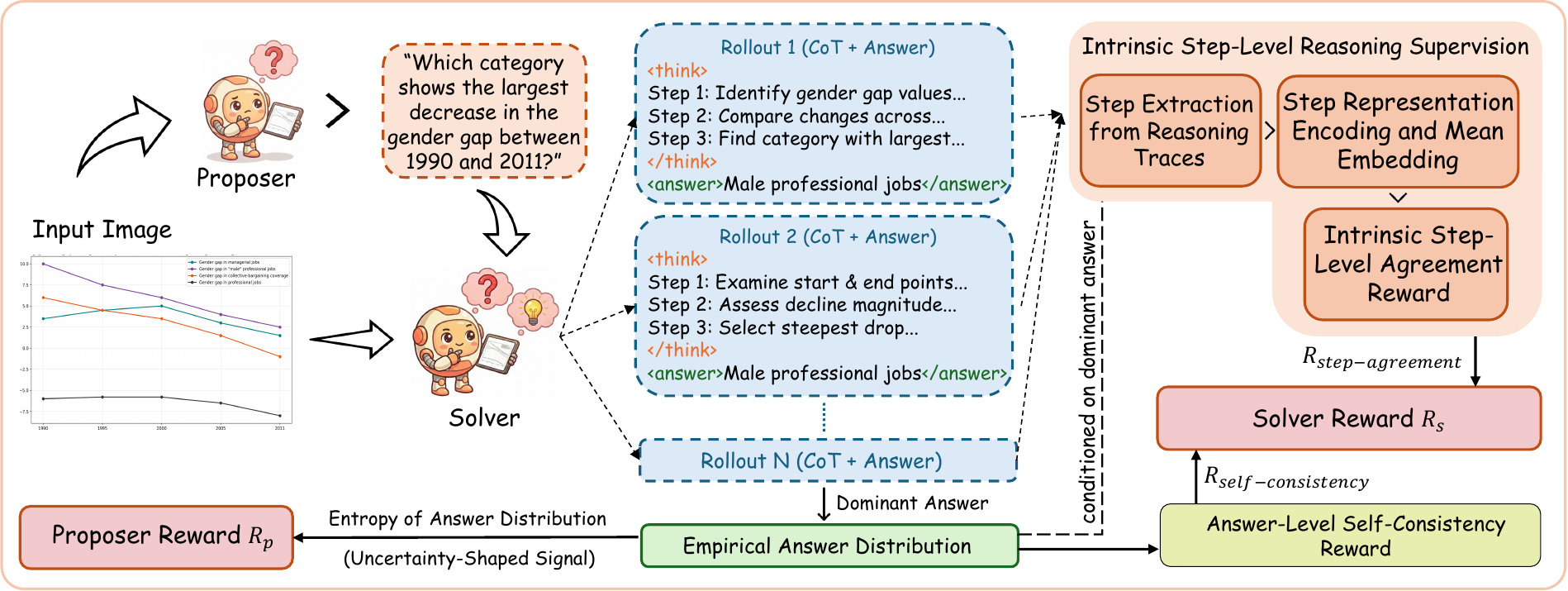}
  \caption{\textbf{Overview of our \textsc{iReasoner} pipeline.}
  From an unlabeled image $x$, a \emph{Proposer} policy $\pi_p$ generates a question $q$, and a \emph{Solver} policy $\pi_s$ produces $N$ sampled reasoning rollouts, inducing an empirical answer distribution $p(a\mid x,q)$.
  The distribution entropy provides an uncertainty-shaped reward for the \emph{Proposer} and selects the dominant answer group for \emph{Solver}-side step supervision.
  The \emph{Solver} reward combines answer-level self-consistency with an intrinsic step-level agreement signal computed over intermediate reasoning traces by (i) extracting numbered steps, (ii) embedding each step, and (iii) forming per-step prototypes within the dominant-answer group.
  This yields intrinsic step-level supervision without annotated question--answer pairs or external verifiers.}
  \label{fig:pipeline}
\end{figure*}

To summarize, our contributions are as follows:
\begin{itemize}
    \item We introduce \textsc{iReasoner}, a fully unsupervised self-evolving framework that brings intermediate reasoning into the optimization loop for \emph{Proposer}--\emph{Solver} self-evolution on unlabeled images.

    \item We propose an intrinsic CoT agreement reward that scores step-level alignment among \emph{Solver} rollouts, providing trajectory-aware supervision that distinguishes reasoning paths leading to the same answer without labeled data or external judges.

    \item Starting from Qwen2.5-VL-7B, we empirically demonstrate that rewarding CoT improves self-evolving LMMs across diverse multimodal reasoning benchmarks, yielding gains of up to $+2.1$ points under fully unsupervised post-training.
\end{itemize}

\section{Related Works}

\paragraph{Self-Evolution in LMMs. }
In fully unsupervised image-only settings, EvoLMM~\cite{thawakar2025evolmmselfevolvinglargemultimodal} instantiates a cooperative \emph{Proposer}–\emph{Solver} trained with continuous self-consistency rewards, while VisPlay~\cite{he2025visplayselfevolvingvisionlanguagemodels} alternates an image-conditioned questioner and a multimodal reasoner using group-relative rewards to balance difficulty and answer quality. Vision-Zero~\cite{wang2025visionzeroscalablevlmselfimprovement} frames learning as strategic visual self-play over games generated from image pairs with RLVR-style updates. Other extensions broaden the loop: C2-Evo~\cite{l1p2} co-evolves synthetic multimodal data to calibrate training challenges, and Agent0-VL~\cite{l1p3} integrates tools for reasoning and self-verification. Vision-SR1 self-rewards a decomposed perception trace but still depends on ground-truth answers \cite{l1p7}. Complementary methods reduce hallucinations via preference-based signals, such as CSR’s visually constrained rewards and SIMA’s in-context visual self-critic \cite{l1p5,l1p6}. Separately, self-refinement pipelines filter synthetic IQA triplets via triangular consistency \cite{l1p9} or bootstrap fine-grained perception through reconstruction and staged RL \cite{l1p10}.

\paragraph{Optimizing Chain-of-Thought for Reasoning. }
A growing body of work explores how to elicit, control, and improve CoT reasoning in LLMs and LMMs. On the prompting and efficiency side, CAR~\cite{l2p1} adaptively routes between short answers and long rationales using uncertainty signals, while token-control methods allocate reasoning budgets based on complexity to reduce overhead~\cite{l2p9}. CoT quality can also be improved through context selection, by retrieving multimodal demonstrations via cross- and intra-modal similarity~\cite{l2p2}. Other approaches focus on the reasoning trajectory itself. Multimodal-CoT~\cite{l2p4} separates rationale generation from answer inference to improve convergence and reduce hallucinations, and R3V~\cite{l2p5} refines multimodal rationales through reflection and self-training. Training-time objectives align intermediate reasoning, including cascaded self-evaluation and filtering~\cite{l2p3}, preference optimization over CoT steps~\cite{l2p7}, and causal analyses that prune redundant steps or insert missing ones~\cite{l2p6}. Complementary studies examine when long CoTs emerge under SFT or RL and propose quantification frameworks such as reasoning boundaries, to guide CoT optimization~\cite{l2p8,l2p10}.

Despite this progress, the two lines above remain only loosely connected. Self-evolving LMM pipelines largely optimize outcome-level signals, while CoT optimization methods typically assume labeled data, external judges, or offline supervision to assess reasoning quality. As a result, intermediate reasoning cannot yet be evaluated or optimized in a purely unsupervised self-evolution loop; \textsc{iReasoner} bridges this gap by turning cross-rollout step agreement into an intrinsic reward that directly trains the reasoning process during self-improvement.

\section{Method}

\begin{figure}[t]
    \centering
    \includegraphics[width=\columnwidth]{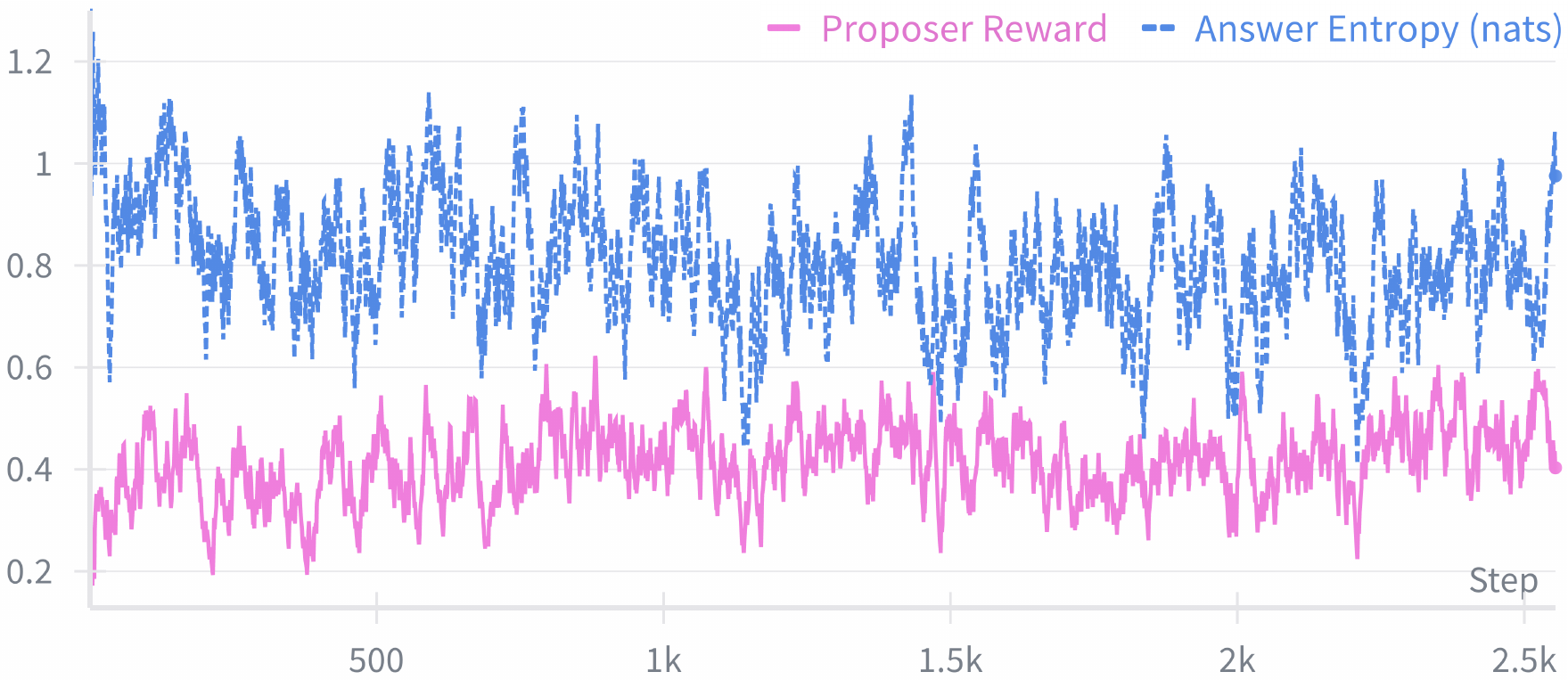}
    \includegraphics[width=\columnwidth]{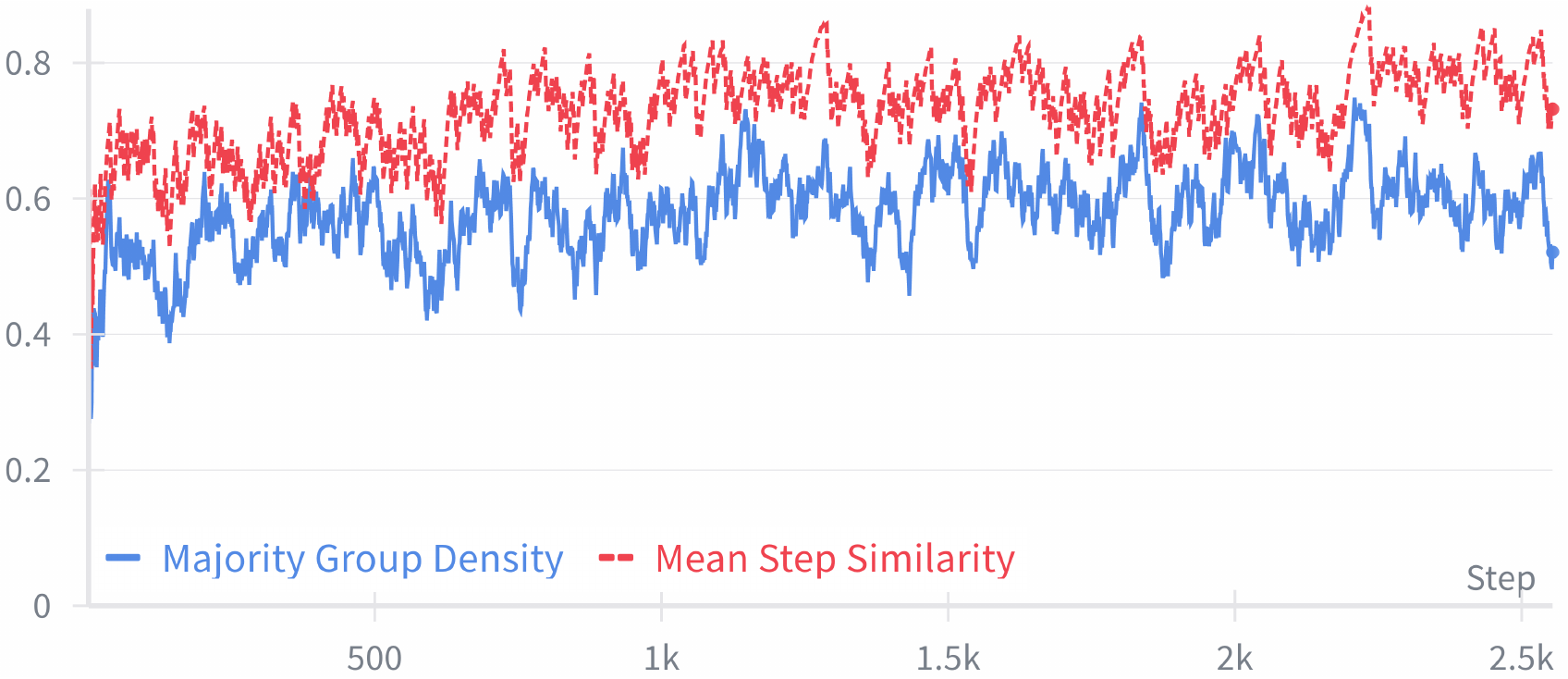}
    \caption{\textbf{Question shaping and \emph{Solver} reasoning behavior over training steps.}
    \textit{(Top)} \emph{Proposer} reward remains stable (around $0.3$--$0.5$) while answer entropy stays in a moderate band (roughly $0.6$--$1.1$ nats), consistent with sustained intermediate difficulty rather than degeneracy.
    \textit{(Bottom)} Majority-group density and mean step similarity increase over training, indicating a larger fraction of \emph{Solver} samples agree on the dominant answer and that their intermediate steps become more aligned.}
    \label{fig:question_reasoning}
\end{figure}

\subsection{Overview}
\label{sec:method-overview}

We introduce \textsc{iReasoner}, a fully unsupervised self-evolving framework that improves an LMM by explicitly optimizing the \emph{Solver}'s CoT, not only its final answers.
Given unlabeled images $x \sim \mathcal{D}$, we adopt the \emph{Proposer}--\emph{Solver} self-evolution regime from prior work: a \emph{Proposer} $\pi_p$ generates a visually grounded question $q \sim \pi_p(\cdot\mid x)$ and a \emph{Solver} $\pi_s$ samples $N$ rollouts $y_i \sim \pi_s(\cdot\mid x,q)$ (Fig.~\ref{fig:pipeline}).
Each \emph{Solver} rollout is structured as
$y_i=\langle \texttt{<think>}~t_i~\texttt{</think><answer>}~a_i~\texttt{</answer>}\rangle$,
where $t_i$ is an explicit multi-step trace and $a_i$ is the extracted answer.
Sampling induces an empirical answer distribution over normalized answers,
\begin{equation}
p(a \mid x,q) \;=\; \frac{1}{N}\sum_{i=1}^{N}\mathbf{1}[a_i=a],
\label{eq:empirical-dist}
\end{equation}
which provides intrinsic supervision without labeled question--answer pairs or external verifiers.

\begin{figure}[t]
  \centering
  \includegraphics[width=\columnwidth]{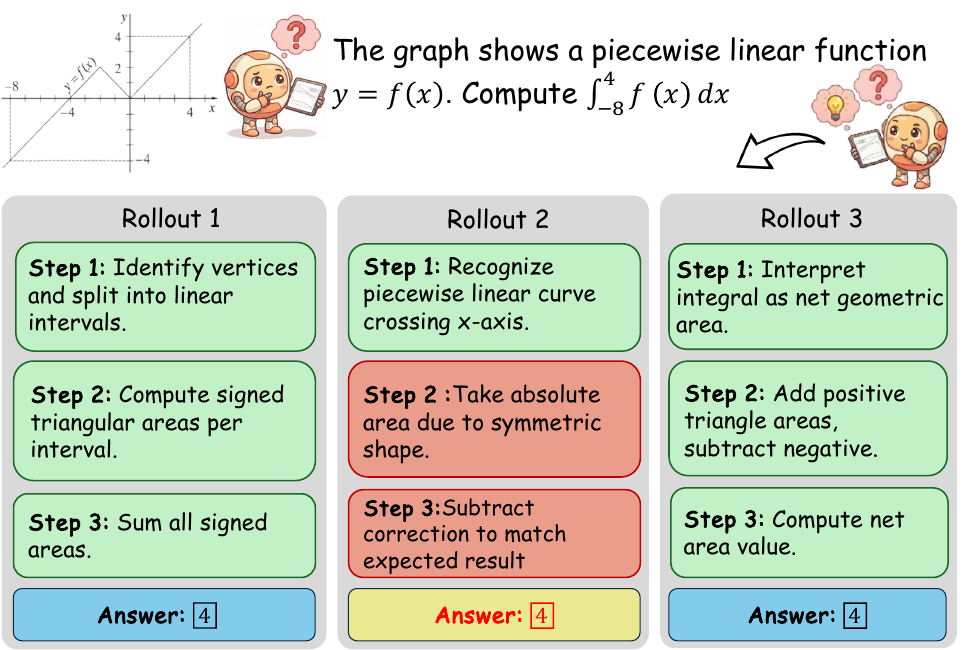}
  \caption{\textbf{Outcome-only self-consistency treats distinct CoTs similarly.}
For the same image-question pair, three \emph{Solver} rollouts produce the same final answer, but their intermediate steps differ: Rollouts 1 and 3 follow a consistent signed-area decomposition, while Rollout 2 deviates via incorrect intermediate claims yet still ends at the same answer. Since outcome-only intrinsic rewards depend only on answer agreement, these rollouts receive nearly identical learning signal despite qualitatively different reasoning traces, motivating step-aware supervision in \textsc{iReasoner}.}
  \label{fig:diff_cots}
\end{figure}

As in EvoLMM~\cite{thawakar2025evolmmselfevolvinglargemultimodal}, outcome-level intrinsic verification is derived from $p(a\mid x,q)$: the \emph{Solver} is rewarded for answer self-consistency, while the \emph{Proposer} is shaped toward intermediate-difficulty questions using the answer entropy
$H(x,q)=-\sum_a p(a\mid x,q)\log p(a\mid x,q)$.
We adopt this \emph{Proposer}-side uncertainty shaping and observe stable \emph{Proposer} reward and sustained non-degenerate answer entropy throughout training (Fig.~\ref{fig:question_reasoning}, top), indicating that the self-evolution loop remains in a meaningful difficulty regime rather than collapsing to trivial questions.

However, outcome-only supervision leaves intermediate reasoning weakly constrained: rollouts that share an answer can receive nearly identical learning signal despite very different CoTs (as illustrated in Fig.~\ref{fig:diff_cots}). 
To this end, \textsc{iReasoner} adds trajectory-aware supervision by an intrinsic mechanism that scores agreement of intermediate steps among rollouts that converge to the same answer, and integrating it into \emph{Solver} optimization (Fig.~\ref{fig:pipeline}).

\subsection{Intrinsic CoT Agreement Reward}
\label{sec:intrinsic-cot}

Our goal is to make intermediate reasoning learnable under unsupervised self-evolution.
For a fixed $(x,q)$, strong solutions should not only agree on the final answer, but also reuse consistent intermediate claims---especially in early grounding steps.
We therefore measure step-wise agreement within rollouts that already agree on the answer.

\begin{figure}[t]
    \centering
    \includegraphics[width=\columnwidth]{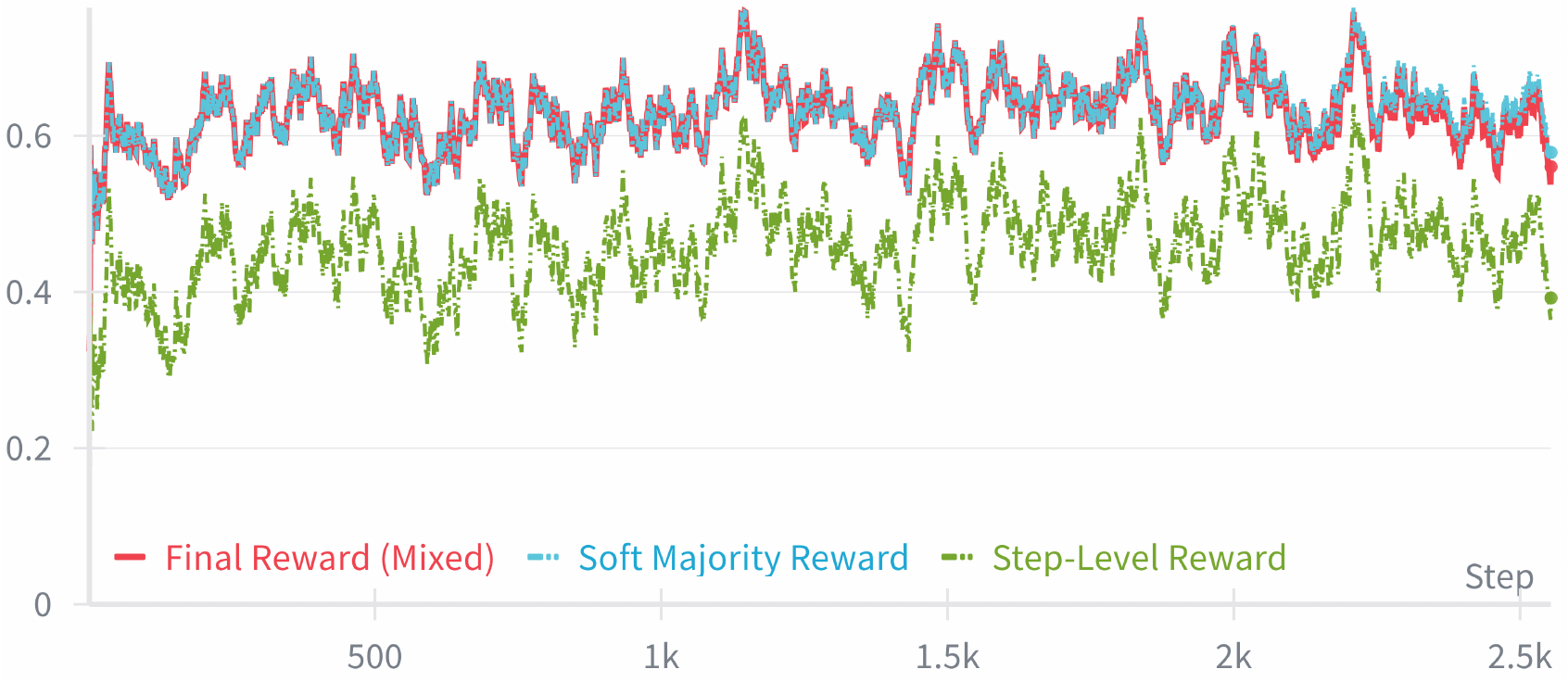}
    \vspace{-0.5em}
    \includegraphics[width=\columnwidth]{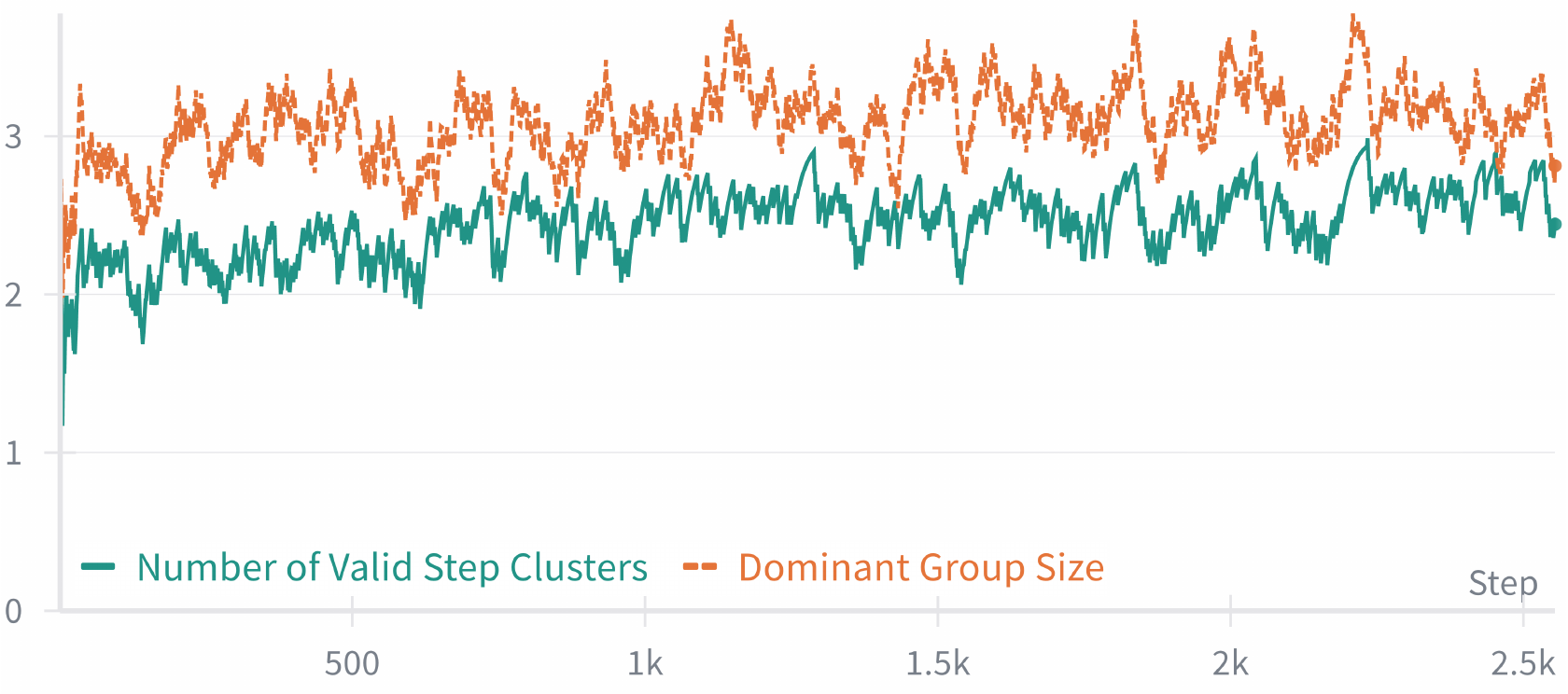}
    \caption{\textbf{Reward components and step-structure trends over training steps.}
    \textit{(Top)} The final mixed \emph{Solver} reward stays in the high range (about $0.6$--$0.7$) while the step-level term provides additional signal beyond soft-majority answer consistency.
    \textit{(Bottom)} The dominant-answer group size and the number of valid step positions used for agreement increase into the $2$--$3$ range, indicating that an increasing portion of the trace is consistently comparable across samples.}
    \label{fig:reward_clustering}
\end{figure}

We view each \emph{Solver} trace as steps $t_i=(s_{i,1},\ldots,s_{i,J_i})$, where $s_{i,j}$ is the $j$-th intermediate claim.
To make agreement meaningful, we elicit a consistent step interface so that the same index $j$ plays a comparable role across rollouts (e.g., early grounding vs.\ later computation).
Let $\hat{a}=\arg\max_a p(a\mid x,q)$ be the dominant answer under Eq.~\eqref{eq:empirical-dist}, and let
\begin{equation}
\mathcal{G}=\{i \mid a_i=\hat{a}\}, \qquad \rho=\left(\frac{|\mathcal{G}|}{N}\right)^{\gamma}, \ \gamma\ge 0
\label{eq:dominant-group}
\end{equation}
denote the dominant-answer group and a density-based reliability factor.
Conditioning on $\mathcal{G}$ anchors step agreement to a stable outcome and avoids reinforcing agreement among off-target rollouts; $\rho$ downweights step supervision when few samples agree.

To compare steps, we embed each step into $e_{i,j}=f(s_{i,j})\in\mathbb{R}^d$ using the model's internal text representations.
In our implementation, $f(\cdot)$ is the $\ell_2$-normalized mean of input-token embeddings for the step text (with a fixed token budget), which provides a lightweight semantic representation aligned with the \emph{Solver}'s language space.
For each index $j$, we form a dominant-group prototype
\begin{equation}
\mu_j \;=\; \frac{1}{|\mathcal{G}_j|}\sum_{i\in \mathcal{G}_j} e_{i,j}, \qquad
\mathcal{G}_j=\{i\in\mathcal{G}\mid j\le J_i\},
\label{eq:prototype}
\end{equation}
and score per-step agreement by cosine similarity $r_{i,j} \;=\; \mathrm{sim}(e_{i,j},\mu_j)\ \in [-1,1].$
Because early steps typically carry grounding and setup, we aggregate with decaying weights $w_1>w_2>\cdots$:
\begin{equation}
\tilde{r}^{\text{step}}_i \;=\; \sum_{j \le J_i} w_j\, r_{i,j}.
\label{eq:raw-step-reward}
\end{equation}
Finally, we define the Intrinsic CoT Agreement Reward as $r^{\text{step}}_i \;=\; \rho \cdot \tilde{r}^{\text{step}}_i.$
Both majority-group density and mean step similarity increase over training (Fig.~\ref{fig:question_reasoning}, bottom), consistent with progressively stabilized \emph{Solver} reasoning traces.

\subsection{Step-Aware Self-Evolution via Reward Integration}
\label{sec:reward-integration}

We train the \emph{Solver} by augmenting outcome-level self-consistency with intrinsic CoT agreement.
Given Eq.~\eqref{eq:empirical-dist}, each rollout receives an answer-level reward
\begin{equation}
r^{\text{ans}}_i \;=\; p(a_i \mid x,q)^{\alpha}\,\bigl(1-\eta\,\bar{\ell}_i\bigr),
\label{eq:ans-reward}
\end{equation}
where $\alpha>0$ controls sharpness, $\bar{\ell}_i\in[0,1]$ is a normalized excess pre-answer length relative to a target budget, and $\eta\ge 0$ sets the length penalty strength.
\textsc{iReasoner} combines Eq.~\eqref{eq:ans-reward} with the step reward:
\begin{equation}
r^{\text{sol}}_i \;=\; (1-\lambda(t))\, r^{\text{ans}}_i \;+\; \lambda(t)\, r^{\text{step}}_i,
\label{eq:Solver-return}
\end{equation}
where $\lambda(t)\in[0,1]$ increases over training with a warmup and ramp.
Early updates rely more on answer-level self-consistency, while later updates place more weight on step agreement once the dominant group and step positions are more reliably populated (Fig.~\ref{fig:reward_clustering}, bottom).
This is also reflected in the reward decomposition (Fig.~\ref{fig:reward_clustering}, top), where the mixed reward remains high while the step-level term provides complementary signal beyond answer-level agreement.

We train both the \emph{Solver} and \emph{Proposer} with KL-regularized policy gradients against a frozen reference policy, which stabilizes self-evolution by limiting drift while preserving reward-driven updates. For the \emph{Solver}, we apply REINFORCE~\cite{williams1992simple} with an EMA baseline $b_s$:
\begin{equation}
\begin{aligned}
\mathcal{L}_s(\theta)
&= -\mathbb{E}_{y\sim \pi_s(\cdot \mid x,q)}
   \!\left[(r^{\text{sol}}(y)-b_s)\,\log \pi_s(y\mid x,q)\right] \\
&\quad + \beta_s\,\mathbb{E}\!\left[
\mathrm{KL}\!\left(
\pi_s(\cdot\mid x,q)\,\|\,\pi_{\text{ref}}(\cdot\mid x,q)
\right)\right],
\end{aligned}
\label{eq:solver-obj}
\end{equation}
The \emph{Proposer} is optimized with the same objective form (baseline $b_p$ and coefficient $\beta_p$), using the entropy-shaped reward $r^p(x,q)=g(H(x,q))$. To maintain a target KL budget, we adapt each coefficient online via a multiplicative controller:
\begin{equation}
\beta \leftarrow \mathrm{clip}\!\left(\beta \cdot \exp\!\Big(\eta\,\frac{\mathrm{KL}-\tau}{\tau}\Big),\ \beta_{\min},\ \beta_{\max}\right),
\label{eq:adaptive_beta}
\end{equation}
which strengthens regularization when the observed KL exceeds $\tau$ and relaxes it otherwise.


\section{Experiments}
\label{sec:experiments}

\subsection{Experimental Setup}
\label{sec:exp-setup}

\begin{table*}[t]
\centering
\small
\setlength{\tabcolsep}{6pt}
\renewcommand{\arraystretch}{1.15}

\newcommand{\best}[1]{\cellcolor{red!20}\textbf{#1}}
\newcommand{\second}[1]{\cellcolor{blue!15}\textbf{#1}}

\resizebox{\textwidth}{!}{%
\begin{tabular}{lcccccccc}
\toprule
& \multicolumn{4}{c}{\textbf{General Visual Understanding}}
& \multicolumn{4}{c}{\textbf{Visual Math}} \\
\cmidrule(lr){2-5}
\cmidrule(lr){6-9}

\rowcolor{gray!12}
\textbf{Model} &
\textbf{InfoGraphic-VQA$_{\textit{val}}$} &
\textbf{AI2D} &
\textbf{ScienceQA} &
\textbf{MMMU$_{\textit{val}}$} &
\textbf{ChartQA} &
\textbf{MathVista} &
\textbf{MathVision} &
\textbf{MathVerse} \\
\midrule

Vision-Zero$^{\dagger}$~\cite{wang2025visionzeroscalablevlmselfimprovement} &
80.35 & 82.64 & 88.50 & 51.44 & 84.24 & 68.43 & 23.96 & 43.86 \\

VisPlay$^{*}$~\cite{he2025visplayselfevolvingvisionlanguagemodels} &
-- & -- & -- & 38.27 & -- & -- & \best{31.15} & 39.14 \\

\hdashline

Qwen2.5-VL-7B (Baseline)~\cite{bai2025qwen25vltechnicalreport} &
80.44 & 82.61 & 88.30 & 51.11 & 84.00 & 68.47 & 23.91 & 43.78 \\

Qwen2.5-VL-7B w/ Discrete Reward~\cite{thawakar2025evolmmselfevolvinglargemultimodal} &
80.52 & 82.18 & 87.98 & 50.84 & 84.62 & 68.88 & 22.52 & 42.10 \\

EvoLMM~\cite{thawakar2025evolmmselfevolvinglargemultimodal} &
\second{81.06} & \second{83.41} & \second{89.50} & \second{52.01} &
\best{86.70} & \best{70.52} & 24.81 & \second{44.88} \\

Qwen2.5-VL-7B w/ Discrete Reward + Step-level Majority &
80.78 & 82.95 & 88.92 & 51.48 &
85.42 & 69.31 & 24.12 & 44.18 \\

\textbf{Qwen2.5-VL-7B w/ Cont. Reward + Step-level Majority (Ours)} &
\best{81.56} & \best{83.89} & \best{89.92} & \best{52.37} &
\second{85.78} & \second{69.74} & \second{25.29} & \best{45.91} \\

\bottomrule
\end{tabular}%
}

\vspace{-0.45em}
\caption{\textbf{Evaluation results across eight multimodal reasoning benchmarks.}
Benchmarks are grouped into general visual understanding and visual mathematics tasks. Best and second-best results are highlighted. Methods marked with ($\dagger$) use external supervision. Methods marked with (*) report results using LLM-as-a-judge evaluation.}
\label{tab:main_results}
\end{table*}

\begin{table*}[t]
\centering
\small
\setlength{\tabcolsep}{5pt}
\renewcommand{\arraystretch}{1.12}

\resizebox{\textwidth}{!}{%
\begin{tabular}{lcccccccc}
\toprule
& \multicolumn{4}{c}{\textbf{General Visual Understanding}}
& \multicolumn{4}{c}{\textbf{Visual Math}} \\
\cmidrule(lr){2-5}
\cmidrule(lr){6-9}

\rowcolor{gray!12}
\textbf{Ablation} &
\textbf{InfoGraphic-VQA$_{\textit{val}}$} &
\textbf{AI2D} &
\textbf{ScienceQA} &
\textbf{MMMU$_{\textit{val}}$} &
\textbf{ChartQA} &
\textbf{MathVista} &
\textbf{MathVision} &
\textbf{MathVerse} \\
\midrule

\rowcolor{gray!8}
Qwen2.5-VL-7B (Baseline) &
80.44 & 82.61 & 88.30 & 51.11 & 84.00 & 68.47 & 23.91 & 43.78 \\

\textbf{Step-level majority (Ours, Full)} &
\textbf{81.56} & \textbf{83.89} & \textbf{89.92} & \textbf{52.37} &
\textbf{85.78} & \textbf{69.74} & \textbf{25.29} & \textbf{45.91} \\

\hdashline
Soft majority reward only (EvoLMM)~\cite{thawakar2025evolmmselfevolvinglargemultimodal} &
81.12 & 83.36 & 89.41 & 51.92 & 86.64 & 70.41 & 24.62 & 44.71 \\

Step-level reward only &
80.61 & 82.69 & 88.44 & 50.98 & 84.38 & 68.73 & 24.18 & 43.87 \\

\hdashline

\rowcolor{gray!8}
\multicolumn{9}{l}{\textit{Step-Level Mechanism Design}} \\
w/o Warmup schedule &
81.04 & 83.21 & 89.26 & 51.74 & 85.02 & 68.97 & 24.63 & 45.11 \\

w/o Position decay &
81.29 & 83.58 & 89.55 & 52.02 & 85.41 & 69.34 & 25.02 & 45.49 \\

w/o Density weighting &
81.18 & 83.46 & 89.47 & 51.88 & 85.29 & 69.19 & 24.91 & 45.32 \\

\hdashline

\rowcolor{gray!8}
\multicolumn{9}{l}{\textit{Reward Shaping Components}} \\
w/o Length penalty &
81.37 & 83.66 & 89.63 & 52.11 & 85.58 & 69.49 & 25.11 & 45.61 \\

Soft majority $\gamma = 0.5$ &
81.09 & 83.31 & 89.34 & 51.71 & 85.21 & 69.07 & 24.72 & 45.12 \\

Soft majority $\gamma = 1.0$ &
80.83 & 83.08 & 89.02 & 51.49 & 84.96 & 68.82 & 24.51 & 44.88 \\

\bottomrule
\end{tabular}%
}

\vspace{-0.45em}
\caption{\textbf{Ablation study of intrinsic reasoning supervision across eight benchmarks.}
The full configuration achieves the strongest overall performance across both task groups. Answer stability alone performs best on highly verifiable benchmarks, while adding step-wise reasoning reward improves transfer on tasks where intermediate structure is informative. Removing mechanism or shaping components leads to consistent regressions, with warmup having the largest effect.}
\label{tab:ablations}
\end{table*}

\paragraph{Training details.}
Our training pool contains 2.5k images sampled from six widely used sources (ChartQA~\cite{chartqa}, AI2D~\cite{ai2d}, InfoGraphic-VQA~\cite{infovqa}, PlotQA~\cite{plotqa}, ChartX~\cite{chartx}, Geometry3K~\cite{geometry3k}). We train on images only; no question--answer pairs, captions, metadata, or external reward models are used at any stage.

We initialize from Qwen2.5-VL-7B-Instruct \cite{bai2025qwen25vltechnicalreport} and train in the \emph{Proposer}--\emph{Solver} self-evolution regime using lightweight LoRA~\cite{hu2022lora} adapters for both roles while keeping the backbone frozen. For each image, the \emph{Proposer} samples one question and the \emph{Solver} samples $N{=}5$ reasoning rollouts. The \emph{Proposer} is updated every 5 iterations. The \emph{Solver} is optimized with KL-regularized REINFORCE using our mixed intrinsic reward (answer stability + step-level agreement), with a warmup schedule that ramps the step component from 0 to a maximum weight of 0.7. We train for 2.5k steps using AdamW (learning rate $10^{-6}$, weight decay 0.01, gradient clipping 1.0) in bfloat16 on 8$\times$ AMD MI250X GPUs; the full run completes in approximately 35 hours.

\paragraph{Evaluation protocol.}
We train \textsc{iReasoner} from the instruction-tuned Qwen2.5-VL-7B model~\cite{bai2025qwen25vltechnicalreport} using an intrinsic CoT-guided RL pipeline in a fully unlabeled setting. Training uses 2{,}500 images from the training splits of ChartQA~\cite{chartqa}, AI2D~\cite{ai2d}, InfoGraphic-VQA~\cite{infovqa}, PlotQA~\cite{plotqa}, ChartX~\cite{chartx} (400 each), and Geometry3K~\cite{geometry3k} (500), with no QA pairs, captions, metadata, or external rewards, preventing ground-truth leakage. Questions are generated from images by a Proposer that self-calibrates difficulty via entropy-based co-evolution, maintaining answer entropy in the 0.6--1.1 nat range.
We evaluate on eight benchmarks: ChartQA~\cite{chartqa}, MathVista~\cite{mathvista}, MathVision~\cite{mathvision}, MathVerse~\cite{mathverse}, InfoGraphic-VQA~\cite{infovqa}, AI2D~\cite{ai2d}, ScienceQA~\cite{scienceqa}, and MMMU~\cite{mmmu}, using official splits and standard accuracy. All models use identical inference settings (no task-specific tuning), so performance reflects self-evolution rather than evaluation-time customization. Evaluations run with \texttt{lmms-eval}~\cite{lmms_eval} on AMD MI250X GPUs in bfloat16.

\subsection{Main Benchmark Results}
Table~\ref{tab:main_results} reports results on eight multimodal reasoning benchmarks spanning general visual understanding and visual mathematics. We make three primary observations.

\begin{figure*}[t]
\centering
\small

\begin{minipage}[t]{0.6\textwidth}
\centering
\setlength{\tabcolsep}{6pt}
\renewcommand{\arraystretch}{1.15}

\resizebox{\textwidth}{!}{%
\begin{tabular}{lcccccccc}
\toprule
& \multicolumn{4}{c}{\textbf{General Visual Understanding}}
& \multicolumn{4}{c}{\textbf{Visual Math}} \\
\cmidrule(lr){2-5}
\cmidrule(lr){6-9}

\rowcolor{gray!12}
\textbf{Max Reasoning Steps} &
\textbf{InfoGraphic-VQA$_{\textit{val}}$} &
\textbf{AI2D} &
\textbf{ScienceQA} &
\textbf{MMMU$_{\textit{val}}$} &
\textbf{ChartQA} &
\textbf{MathVista} &
\textbf{MathVision} &
\textbf{MathVerse} \\
\midrule

4 steps  &
80.92 & 83.12 & 89.21 & 51.82 &
85.10 & 69.08 & 24.71 & 45.02 \\

6 steps  &
81.31 & 83.61 & 89.68 & 52.21 &
85.54 & 69.52 & 25.06 & 45.62 \\

\rowcolor{gray!8}
\textbf{8 steps (default)} &
\textbf{81.56} & \textbf{83.89} & \textbf{89.92} & \textbf{52.37} &
\textbf{85.78} & \textbf{69.74} & \textbf{25.29} & \textbf{45.91} \\

10 steps &
81.42 & 83.74 & 89.79 & 52.26 &
85.63 & 69.61 & 25.12 & 45.71 \\

\bottomrule
\end{tabular}
}
\vspace{-0.45em}
\captionof{table}{\textbf{Sensitivity to the maximum number of reasoning steps.}
We report performance across all eight benchmarks while varying the maximum number of parsed reasoning steps used to extract intermediate reasoning structure. Smaller step budgets truncate useful intermediate information, while overly large budgets can introduce noisy or redundant steps. The default setting of 8 steps used in our main experiments provides a strong balance and achieves robust performance across both general visual understanding and visual mathematics tasks.}
\label{tab:step_budget}
\end{minipage}
\hfill
\begin{minipage}[t]{0.38\textwidth}
\centering
\vspace{-3em}
\includegraphics[width=\textwidth]{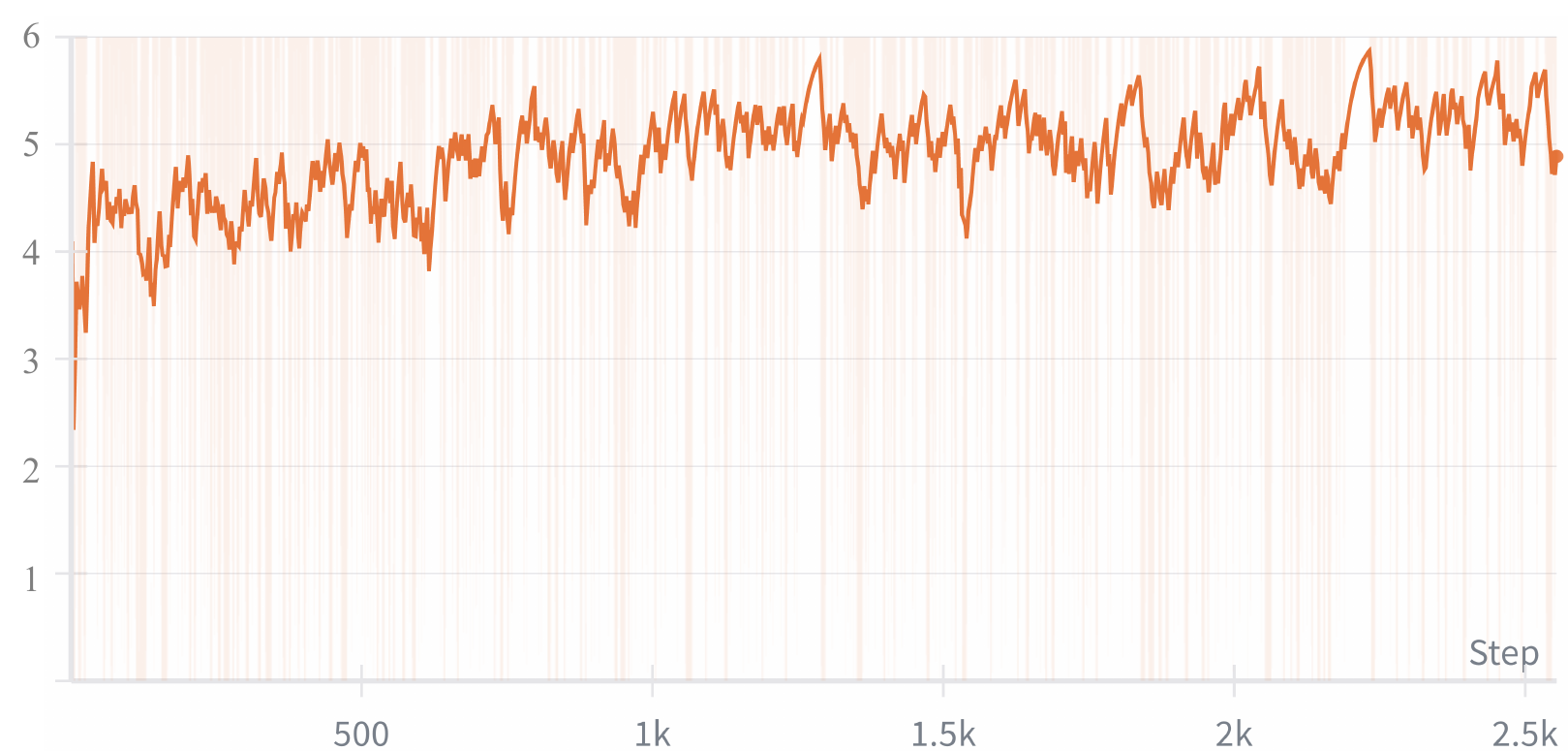}

\vspace{-0.45em}
\captionof{figure}{\textbf{Evolution of reasoning step usage over training.}
Average number of extracted reasoning steps per solution (capped at eight), which gradually stabilizes over training, reflecting more consistent and structured intermediate reasoning.}
\label{fig:step_evolution}
\end{minipage}

\end{figure*}

\paragraph{\textsc{iReasoner} consistently improves over the seed model across both task groups.}
Across general visual understanding benchmarks, \textsc{iReasoner} improves over Qwen2.5-VL-7B by +1.12 on InfoGraphic-VQA, +1.28 on AI2D, +1.62 on ScienceQA, and +1.26 on MMMU, corresponding to an average gain of +1.32. On visual mathematics benchmarks, \textsc{iReasoner} improves on all four datasets (+1.64 on average), with the largest gain on MathVerse (+2.13) and a consistent improvement on MathVision (+1.38).

\paragraph{Step-wise reasoning reward improves general-purpose transfer beyond answer-level agreement.}
Compared to EvoLMM~\cite{thawakar2025evolmmselfevolvinglargemultimodal}, \textsc{iReasoner} improves all four general benchmarks: InfoGraphic-VQA (+0.50), AI2D (+0.48), ScienceQA (+0.42), and MMMU (+0.36). This pattern is strongest on tasks where multiple intermediate reasoning traces can reach the same final answer (such as those in MMMU and MathVerse), showing that cross-trajectory step agreement provides supervision that is not captured by answer-level consistency alone. On visual math benchmarks, \textsc{iReasoner} improves MathVision (24.81$\rightarrow$25.29) and MathVerse (44.88$\rightarrow$45.91), while EvoLMM remains stronger on ChartQA and MathVista. This contrast is consistent with answer-stability rewards being particularly effective on highly verifiable short-answer settings, whereas step-level alignment contributes more when intermediate structure is informative and transferable.

\paragraph{Step-level agreement requires continuous rewards to be effective.}
To isolate the interaction between reward type and step-level supervision, we include a discrete-reward variant augmented with step-level agreement. While adding step-level agreement to a discrete reward improves over the discrete-only baseline (e.g., InfoGraphic-VQA 80.52$\rightarrow$80.78; MathVerse 42.10$\rightarrow$44.18), the gains remain limited and do not match those achieved by \textsc{iReasoner}. In contrast, combining step-level agreement with continuous intrinsic rewards yields consistent improvements across all eight benchmarks, indicating that step-level signals benefit from smoother credit assignment, where partial alignment in intermediate steps can be rewarded even when answers are not perfectly consistent.

\paragraph{Comparison to prior self-evolving LMMs.}
We also compare against other self-evolving LMMs under their reported evaluation protocols. VisPlay~\cite{he2025visplayselfevolvingvisionlanguagemodels}, which operates in a similar unsupervised self-evolving regime, achieves higher performance on MathVision, while \textsc{iReasoner} substantially outperforms VisPlay on MMMU$_{\textit{val}}$ (52.37 vs.\ 38.27) and MathVerse (45.91 vs.\ 39.14). Vision-Zero~\cite{wang2025visionzeroscalablevlmselfimprovement} is included for context but relies on external supervision and is therefore not directly comparable.

\newlength{\panelH}
\setlength{\panelH}{0.3\textheight} 

\begin{figure*}[t]
\centering

\begin{minipage}[t]{0.65\textwidth}
  \centering
  \vspace{0pt}
  \includegraphics[width=\linewidth,height=\panelH,keepaspectratio]{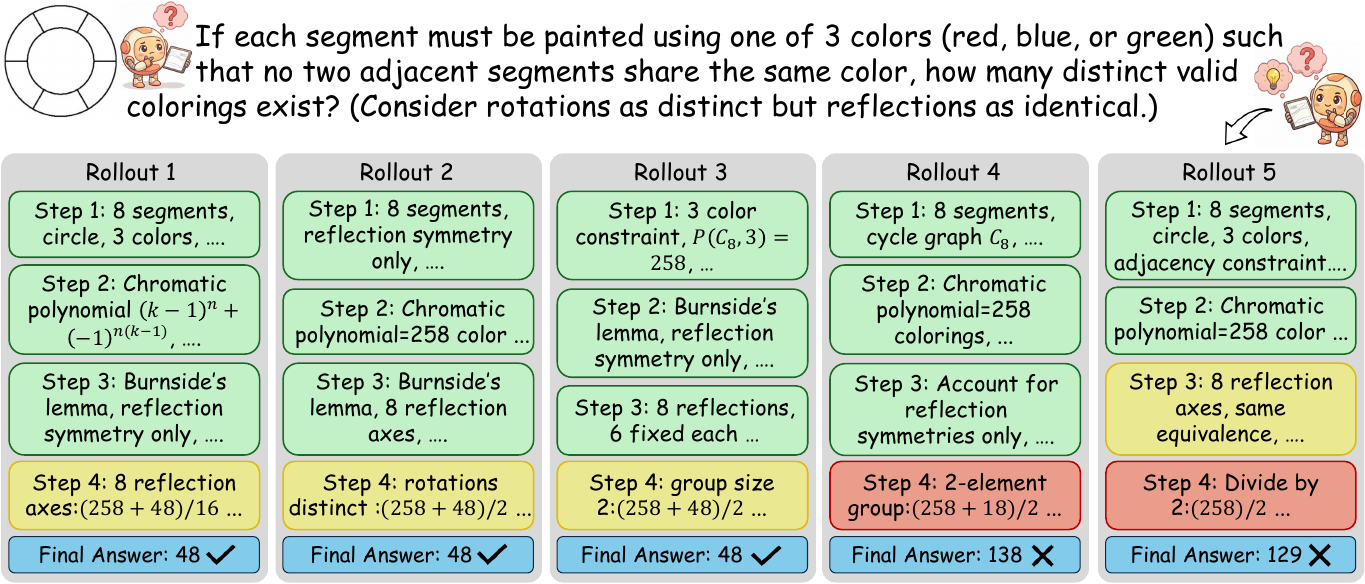}
\end{minipage}\hfill
\begin{minipage}[t]{0.32\textwidth}
  \centering
  \vspace{0pt}
  \includegraphics[width=\linewidth,height=0.49\panelH,keepaspectratio]{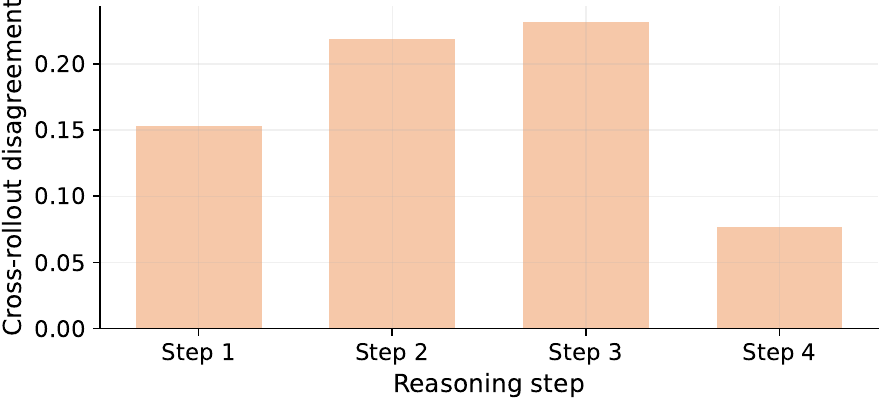}


  \includegraphics[width=\linewidth,height=0.49\panelH,keepaspectratio]{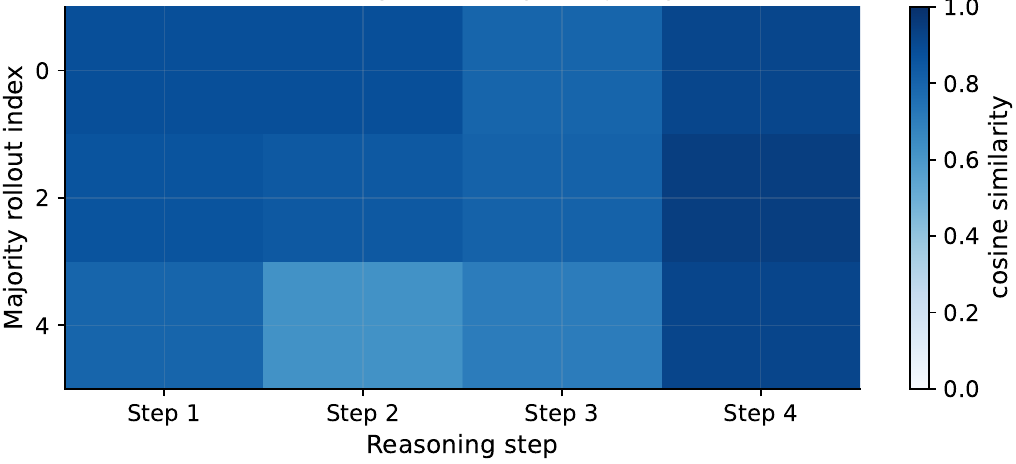}
\end{minipage}
\vspace{-0.45em}
\caption{\textbf{Majority-group rollouts and step-level agreement diagnostics.}
Left: qualitative sample of an image, \emph{Propposer}-generated question and five \emph{Solver} rollouts with four reasoning steps.
Right (top): per-step disagreement $D_j$ computed from leave-one-out similarities.
Right (bottom): leave-one-out step similarity heatmap $S_{i,j}$ over rollouts $i\in\mathcal{G}$ and step indices $j$ (higher indicates greater consistency with the group).}
\label{fig:loo_and_divergence_triptych}
\end{figure*}

\subsection{Ablation Study}
Table~\ref{tab:ablations} ablates the two intrinsic rewards and key design choices in cross-trajectory step alignment.

\noindent\textbf{Answer stability is strongest on highly verifiable benchmarks, while step-wise reward improves transfer.}
Using only the answer-stability reward yields large gains on ChartQA and MathVista, improving over the seed model by +2.64 (84.00$\rightarrow$86.64) and +1.94 (68.47$\rightarrow$70.41). Adding step-wise reward improves benchmarks where intermediate structure matters: relative to answer stability alone, \textsc{iReasoner} improves InfoGraphic-VQA by +0.44, AI2D by +0.53, ScienceQA by +0.51, MMMU by +0.45, and also increases MathVision by +0.67 and MathVerse by +1.20. At the same time, it is slightly lower on ChartQA (-0.86) and MathVista (-0.67), consistent with a trade-off between answer-level agreement and enforcing step-level structure on highly verifiable tasks.

\noindent\textbf{Step-wise reward alone is weak, but becomes effective with answer stability.}
Using only step-wise reward yields small gains over the seed model (e.g., +0.17 on InfoGraphic-VQA; +0.09 on MathVerse) and does not improve MMMU. In contrast, combining both signals substantially improves over step-wise reward alone (e.g., +1.48 on ScienceQA; +2.04 on MathVerse). Overall, cross-trajectory step alignment benefits from an answer-level signal that reduces reward noise early in training.

\noindent\textbf{Mechanism and shaping choices matter, and warmup is most important.}
Removing the warmup schedule produces the largest and most uniform regression (average -0.68 across benchmarks), including -0.76 on ChartQA and -0.80 on MathVerse. Removing position decay or density weighting also reduces performance with smaller but consistent drops. Reward shaping has systematic effects: removing the length penalty reduces MathVerse by -0.30, while removing shaping entirely ($\gamma{=}1.0$) leads to larger degradations (e.g., -0.90 on ScienceQA and -1.03 on MathVerse).

\begin{table*}[t]
\centering
\caption{%
  \textbf{Top:} Comparison of baseline models and their \textsc{iReasoner}-enhanced counterparts across three scales on eight benchmarks.
  \textbf{Bottom:} Inference-time CoT ablation. Suppressing chain-of-thought at inference shows that step-level training improves the model's underlying representations (row 2 vs.\ row 1), and that the aligned reasoning chains contribute active inferential work at test time (row 3 vs.\ row 2).%
}
\label{tab:combined}
\resizebox{\textwidth}{!}{%
\begin{tabular}{lcccccccc}
\toprule
\textbf{Model} & \textbf{InfoGraphic-VQA} & \textbf{AI2D} & \textbf{ScienceQA} & \textbf{MMMU} & \textbf{ChartQA} & \textbf{MathVista} & \textbf{MathVision} & \textbf{MathVerse} \\
\midrule
Qwen2.5-VL-3B (Baseline)           & 78.00 & 79.80 & 86.40 & 47.20 & 80.10 & 65.10 & 19.40 & 39.20 \\
\textbf{Qwen2.5-VL-3B + iReasoner} & \textbf{78.81} & \textbf{80.11} & \textbf{86.97} & \textbf{47.89} & \textbf{81.67} & \textbf{66.28} & \textbf{20.90} & \textbf{39.64} \\
\midrule
Qwen2.5-VL-7B (Baseline)           & 80.44 & 82.61 & 88.30 & 51.11 & 84.00 & 68.47 & 23.91 & 43.78 \\
\textbf{Qwen2.5-VL-7B + iReasoner} & \textbf{81.56} & \textbf{83.89} & \textbf{89.92} & \textbf{52.37} & \textbf{85.78} & \textbf{69.74} & \textbf{25.29} & \textbf{45.91} \\
\midrule
Qwen2.5-VL-32B (Baseline)          & 82.80 & 85.30 & 91.60 & 58.10 & 86.90 & 71.90 & 29.80 & 48.90 \\
\textbf{Qwen2.5-VL-32B + iReasoner}& \textbf{83.58} & \textbf{86.27} & \textbf{92.79} & \textbf{59.89} & \textbf{89.97} & \textbf{74.21} & \textbf{31.77} & \textbf{50.42} \\
\midrule\midrule
\multicolumn{9}{l}{\textit{Inference-time CoT Ablation (Qwen2.5-VL-7B scale)}} \\
\midrule
Qwen2.5-VL-7B (Baseline)           & 80.44 & 82.61 & 88.30 & 51.11 & 84.00 & 68.47 & 23.91 & 43.78 \\
iReasoner w/o CoT at inference      & 80.84 & 82.93 & 88.77 & 51.76 & 84.69 & 68.73 & 24.31 & 44.52 \\
\textbf{iReasoner (Full)}           & \textbf{81.56} & \textbf{83.89} & \textbf{89.92} & \textbf{52.37} & \textbf{85.78} & \textbf{69.74} & \textbf{25.29} & \textbf{45.91} \\
\bottomrule
\end{tabular}%
}
\end{table*}

\begin{table}[t]
\centering
\caption{Fraction of training steps where the dominant-answer group matches ground truth.}
\label{tab:correct_majority}
\vspace{-2mm}
\resizebox{\columnwidth}{!}{%
\begin{tabular}{lcc}
\toprule
\textbf{Training Phase} & \textbf{Steps} & \textbf{Correct-Majority Rate} \\
\midrule
Early & 0--1{,}000     & $\sim$76\% \\
Mid   & 1{,}000--2{,}000 & $\sim$86\% \\
Late  & 2{,}000--2{,}500 & $\sim$93\% \\
\bottomrule
\end{tabular}%
}
\end{table}

\subsection{Training Dynamics and Reasoning-Trace Analysis}

\paragraph{Sensitivity to the step budget.}
We analyze sensitivity to the maximum number of intermediate reasoning steps used for cross-trajectory alignment. Table~\ref{tab:step_budget} shows that increasing the step budget from 4 to 8 yields consistent gains (e.g., MathVerse 45.02$\rightarrow$45.91; AI2D 83.12$\rightarrow$83.89), while increasing beyond 8 gives diminishing returns and can slightly regress, consistent with longer traces adding redundant or noisier steps for alignment. Figure~\ref{fig:step_evolution} further shows that the average number of extracted steps per rollout (capped at eight) stabilizes over training, suggesting that intrinsic reasoning supervision promotes more consistent trace structure rather than unbounded growth.

\paragraph{Within-mode divergence under answer agreement.}
Even when answers agree, intermediate reasoning can vary substantially, as rollouts in the dominant-answer group $\mathcal{G}$ may share the same final answer while diverging on key steps (Fig.~\ref{fig:loo_and_divergence_triptych} (left)). To localize where this occurs, we measure step agreement within $\mathcal{G}$ using the same step parsing and lightweight text embeddings as our step-level reward. For each rollout $i\in\mathcal{G}$ and step index $j$, we compute a leave-one-out similarity
$S_{i,j}=\cos(e_{i,j},\mu_{-i,j})$,
where $\mu_{-i,j}$ is the mean step-$j$ embedding over the other majority rollouts. Fig.~\ref{fig:loo_and_divergence_triptych} (bottom right) visualizes $S_{i,j}$ and shows that even under answer agreement, some rollouts deviate sharply at specific step indices while remaining aligned elsewhere. Aggregating across rollouts gives a depth-wise disagreement profile
$D_j = 1-\frac{1}{|\mathcal{G}|}\sum_{i\in\mathcal{G}} S_{i,j}$.
As shown in Fig.~\ref{fig:loo_and_divergence_triptych} (top right), disagreement concentrates in the middle of the trace (steps 2--3) and is lower in later steps, supporting that outcome-only rewards cannot distinguish stable from unstable reasoning within the dominant answer mode, whereas step-level objectives directly target trace parts where rollouts drift.

\subsection{Scaling Across Model Sizes}
\label{sec:appendix_scale}

To assess generality beyond a single model size, we ran \textsc{iReasoner} on Qwen2.5-VL-3B, 7B, and 32B. Table~\ref{tab:combined} (top) reports results across all eight benchmarks. Gains are consistent at every scale. Smaller models (3B, 7B) converge well within approximately 2,500 steps, while the 32B model benefits from a slightly longer run of around 3,000 steps, likely because the larger parameter space needs more gradient updates to calibrate. The core finding holds regardless of size: trajectory-aware step supervision provides reliable additive gains on top of the base model.

\subsection{Reward Reliability \& CoT Faithfulness}
\label{sec:appendix_intrinsic_cot}

A key concern with intrinsic rewards is that a confidently wrong dominant group could reinforce incorrect reasoning. To assess this, we track whether the dominant-answer group matches ground truth at each training phase (used only diagnostically, never during training). As shown in Table~\ref{tab:correct_majority}, the dominant group is already correct roughly three-quarters of the time early in training and improves steadily with \emph{Proposer}--\emph{Solver} co-evolution. This risk is further mitigated by two design choices: a warmup schedule that delays step-level supervision until answer stability improves, and a density weighting factor $\rho$ that downweights step signals when the dominant group is small which is precisely when error risk is highest. To verify that gains reflect genuine reasoning rather than consistent-but-wrong convergence, we log all questions the \emph{Solver} fails during training and re-evaluate them with the final model, which answers 87\% correctly. This indicates real improvement in reasoning ability; future work could incorporate ensemble disagreement for additional robustness.

We also examine whether aligned CoT traces are causally useful or merely post-hoc rationalizations by evaluating the \textsc{iReasoner}-trained model with CoT suppressed at inference (i.e., direct answers without intermediate reasoning). As shown in Table~\ref{tab:combined} (bottom), (1) the model still outperforms the baseline without CoT, indicating that step-level training improves internal representations beyond output formatting; and (2) a consistent gap between no-CoT and full-CoT settings across all eight benchmarks shows that aligned chains contribute meaningfully to reasoning at test time rather than serving as superficial explanations.

\section{Conclusion}

We presented \textsc{iReasoner}, a self-evolving post-training framework for LMMs that brings intermediate reasoning into the optimization loop in a fully unlabeled setting. \textsc{iReasoner} introduces a trajectory-aware intrinsic signal that aligns \emph{Solver} traces across responses that converge to the same answer, addressing a key limitation of outcome-only self-consistency where identical answers can arise from unstable step sequences. Starting from Qwen2.5-VL-7B and training exclusively on unlabeled images, \textsc{iReasoner} yields consistent gains across diverse multimodal reasoning benchmarks and shows that step-aware intrinsic supervision can improve transfer beyond answer-level agreement alone. We hope this work serves as a useful starting point for reasoning-aware self-improvement and more open-ended self-generated curricula under minimal supervision in multimodal systems.

\section*{Limitations}

\textsc{iReasoner} uses only intrinsic signals derived from the model's own samples. As a result, it cannot directly optimize for external correctness: when the dominant-answer group is confidently wrong (e.g., due to early-training noise or perception failures), step-level agreement may reinforce internally consistent but incorrect reasoning. Mitigating this failure mode without introducing external judges remains open.
Our work is also limited in scale and coverage. We train for 2.5k self-evolution steps on 2.5k unlabeled images and report results from a single backbone. Longer runs, larger and more diverse unlabeled image streams, and additional model families are needed to better characterize stability and scaling.
Finally, our training procedure assumes access to model internals to compute log-probability objectives and KL regularization against a reference policy. This makes the approach most applicable to open-weight models and less directly transferable to black-box systems that do not expose token-level likelihoods.

\section*{Acknowledgements}
This work was supported by KAKENHI grant number JP24K15011.

\bibliography{custom}




\end{document}